# Unsupervised Image-to-Image Translation with Generative Adversarial Networks

Hao Dong
Imperial College London
hao.dong11@imperial.ac.uk

Paarth Neekhara
Indian Institute of Technology
pniitucs@iitr.ac.in

Chao Wu
Imperial College London
chao.wu@imperial.ac.uk

Yike Guo
Imperial College London
y.guo@imperial.ac.uk

## Abstract

*It's useful to automatically transform an image from its original form to some synthetic form (style, partial contents, etc.), while keeping the original structure or semantics. We define this requirement as the "image-to-image translation" problem, and propose a general approach to achieve it, based on deep convolutional and conditional generative adversarial networks (GANs), which has gained a phenomenal success to learn mapping images from noise input since 2014. In this work, we develop a two step (unsupervised) learning method to translate images between different domains by using unlabeled images without specifying any correspondence between them, so that to avoid the cost of acquiring labeled data. Compared with prior works, we demonstrated the capacity of generality in our model, by which variance of translations can be conduct by a single type of model. Such capability is desirable in applications like bidirectional translation*

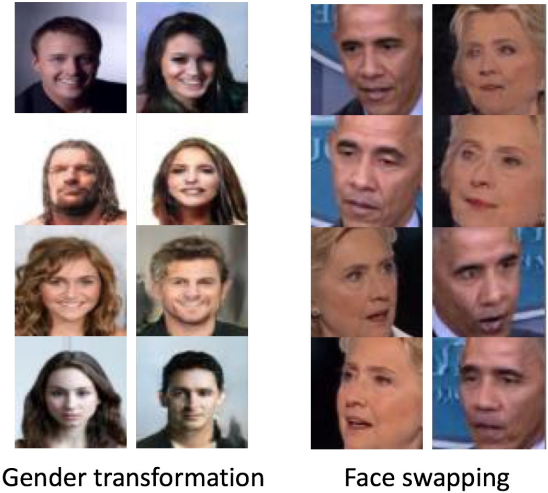

Figure 1. Example results of unsupervised image-to-image translation. The odd columns are input images, the even columns are the corresponding synthesized output images.

## 1. Introduction

In this work we are interested in the problem of translating images from one domain to the images of other domains *e.g.* face of a person to another, aerial image to map, edges image to photos, black and white image to color image. Such translation is useful in creating synthetic images for various purpose. For example, by translating real images to synthetic counterparts, we can easily create balanced sizes of data for image classification tasks. We might also use this method to modify the context of photos (such as weather and background).

The community has been put a lot effort in this direction *e.g.* face swapping [9], changing time of day for outdoor image [19], changing weather of outdoor image [10], composed photo from sketch [3], but each of these tasks has been tackled with separate, special-purpose machinery. It would be more preferable if there is an universal model to achieve all these functions.

Fortunately, based on Generative Adversarial Networks (GANs) [5], a study [7] develop a common framework suitable for different problems, the generator in [7] is an encoder-decoder network which tries to synthesize fake image conditioned on a given image to fool discriminator, while the discriminator tries to identify the fake image by comparing with the corresponding target image. However, this work is supervised learning, therefore, it needs intensive input of labeled data for training, which is not always applicable in real cases.

Recently, another work [20] demonstrated an unsupervised approach for image-to-image translation, the training images do not need to have a specific input and target. It employed a Domain Transfer Network to learn a generative network to map images from source domain to images in target domain. It also adopted a function from both domain to some metric, which is invariant under $G$. The method is general in its scope and does not rely on a predefined family of perceptual losses. Therefore, it can also be used for unsupervised domain adaption [2, 4].

In this work we purposed a two step learning method that utilize conditional GAN to learn the global feature of different domains, and learn to synthesize plausible image from random noise and given class/domain label. After that, a deep convolutional encoder is used to learn a mapping from an image to it's latent representation (noise). We demonstrated that using the trained deep convolutional encoder and conditional generator, images of different domains can be translated bidirectionally. Besides unsupervised learning, our method proposed a new way of training image encoder. To encourage the community to train different dataset, we will make code publicly available at [1].

One significant feature of our model is that it has preliminarily capability of one learning algorithm. According to one learning algorithm hypothesis, popularized by Jeff Hawkins [6], a universal learning machine is a powerful and general model for intelligent agents like human, in which the same network structure can serve for different learning purposes. Such capability is an desirable extension of deep learning algorithms, so that a universal algorithm as along with its deployment can adapt to the various tasks on hand. Compared to prior works in images translation, our method has such universal feature to support different learning scenarios

## 2. Background

In this section, we discuss the related work and the techniques be used in our work. A generative adversarial network (GAN) consists of a generator $G$ and a discriminator $D$ [5]. The discriminator is trained to maximize the probability of assigning the correct label to both training samples and samples generated from generator. While, the generator is trained to minimize the probability of assigning a samples from generator to be fake, so as to generate plausible samples conditioned on the random noise vector $z$. Following this, deep convolutional GAN [17] learns to synthesize images from unlabeled data, then an image can be represented by a fixed number of latent variables, and each of variable reflects a certain feature [17] *e.g.* controlling the window size of bathroom, smile of face and so on. The loss functions of $D$ and $G$ are shown as below, where $P(s|X)$ indicates the score of a image to be real and $X_{fake} = G(z)$.

$$\mathcal{L}_D = log(P(s|D(X_{real}))) + log(1 - P(s|D(X_{fake})))$$
$$\mathcal{L}_G = log(P(s|D(X_{fake}))) \qquad (1)$$

Following this, mass of studies show that images can be synthesized based on conditions such as sentence [18], class label [13], image inpainting [16] and image manipulation by constraints [21] etc. To synthesize image conditioned on a given class label $c$, auxiliary classifier GAN [15] demonstrate how to synthesize high quality image by using a new structure and loss function. In the generator $G$, input the class label $c$ and the noise vector $z$, then a deconvolutional network [17] performs image generation conditioned on both class label $c$ and the noise vector $z$. The discriminator is a convolutional network, which outputs the class label for the training data, that except requiring generate plausible image, the generator also trained to maximize the log-likelihood of the synthesize image to the correct class [14]. The loss functions as below, where $X_{fake} = G(z, c)$:

$$\mathcal{L}_D = log(P(s|D(X_{real}))) + log(1 - P(s|D(X_{fake})))$$
$$+ log(P(c|D(X_{real})))$$
$$\mathcal{L}_G = log(P(s|D(X_{fake}))) + log(P(c|D(X_{fake}))) \qquad (2)$$

To synthesize an image from another image, the input image need to be projected into latent representation. In [21], author introduced a method to train a feedforward network for projecting an image to latent variables. The loss consists of two parts, mean-square-error (MSE) of input and output image for pixel-wise level reconstruction and the loss of hidden outputs for feature level reconstruction. Compared with previous works, we introduced a new method to train an image encoder specially for generator.

The one learning algorithm is a concept in machine learning and neuroscience that has been studied extensively by Je Hawkins [6]. It's based on a memory-prediction framework. The framework is based upon the work on columnar organization of the cerebral cortex which shows that all regions of the cortex perform the same operation.

## 3. Method

Our image translation model is constituted with both deep convolutional encoder and decoder. It has input of images from source domain and output images of target domain(s), which conditioned on a given class/domain label. The training process has two steps: first step for image generator $G$ for all domains, and second step for image encoder $E$ for all domains as well. During the translation process, the conditional class label is applied to input image from source domain and go through the trained network for the output image in target domain.

---
[1] https://github.com/coming-soon

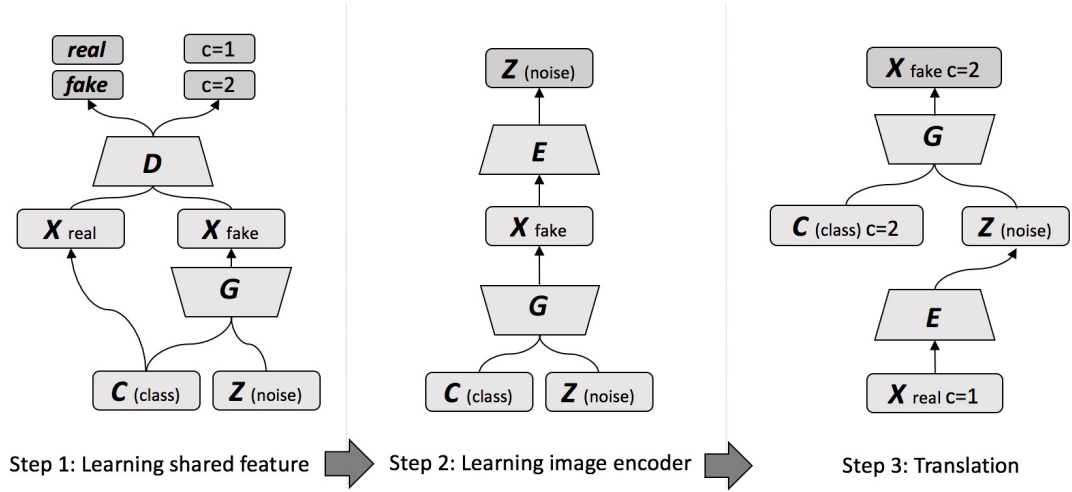

Figure 2. Network architectures of two step learning. This example is for two classes/domains, but our method can be extend to multiple domains.

### 3.1. Learning shared feature

To utilize both the capability of unsupervised feature learning of GANs, and the representation capability of convolutional encoder, we separated the learning process into two steps.

In first step, as the left hand side of Figures 2 shows, we used auxiliary classifier GAN [15] to learn the global shared features of the images sampled from different domains. The shared features are represented as a latent variable (noise) vector $z$. In order to limit the variables into a specific range, we defined it as $z \in \mathbb{R} \sim \mathcal{U}(-1, 1)$.

The idea is based on the fact that class-independent information contains global structure about the synthesized image [15]. Therefore, if the domains are semantically same, or similar (like different faces of peoples, different weather of environment), we expect the common features across the domains (like smile, face pose, background etc) to be captured by the same latent representation in the domains. We leverage this feature for translating the global structure from one domain to the other.

Conditional GANs usually use an embedding layer [18], we found same performance between embedding layer and one-hot format input. Although through experiment, we found the embedding layer contributed less to the performance. We believe this is due to the limit size of sampling domains in our experiment. If there are increasing size of domains (along with their relationships as word2vec [12]), such embedding layer can contribute more to the performance, compared with one-hot format.

### 3.2. Learning image encoder

After the first step, the generator is able to generate corresponding images for different domains by keeping the latent variables fixed and changing the class label. In second step, we learn a mapping from image to global latent variable representation $z$.

Instead of applying trained generator $G$ on top of image encoder $E$ [21] and except that the output image looks the same with the input image, we introduce a new method that applies $E$ after the $G$ from the first step as the middle of Figure 2 shows. We then except the network outputs the same variables with the input variables. Mean-square-error (MSE) between input latent variables (noise) and output variable is used as the loss. We found this method is not only able to reconstruct the detailed feature but it also speed up the training.

The reason of this method working well is that: as the trained generator is able to synthesize image from arbitrary noise, we use the infinite synthesize images from the trained generator instead of the finite real image from dataset, so that we can have much more training data. Such feature is crucial since to train a image encoder with similar level of representation capacity as generator, the training datasets used should be at similar scale.

Besides, there are two keys points about training: firstly, instead of passing more than just one class of images, we passed images of all classes, specifying the class along with it, this enables the image encoder which is able to work for all classes of image so as to achieve bidirectional translation; Secondly, as the noise $z$ has uniform distribution, the outputs of $E$ are scaled into $[-1, 1]$ by $tanh$ activation function.

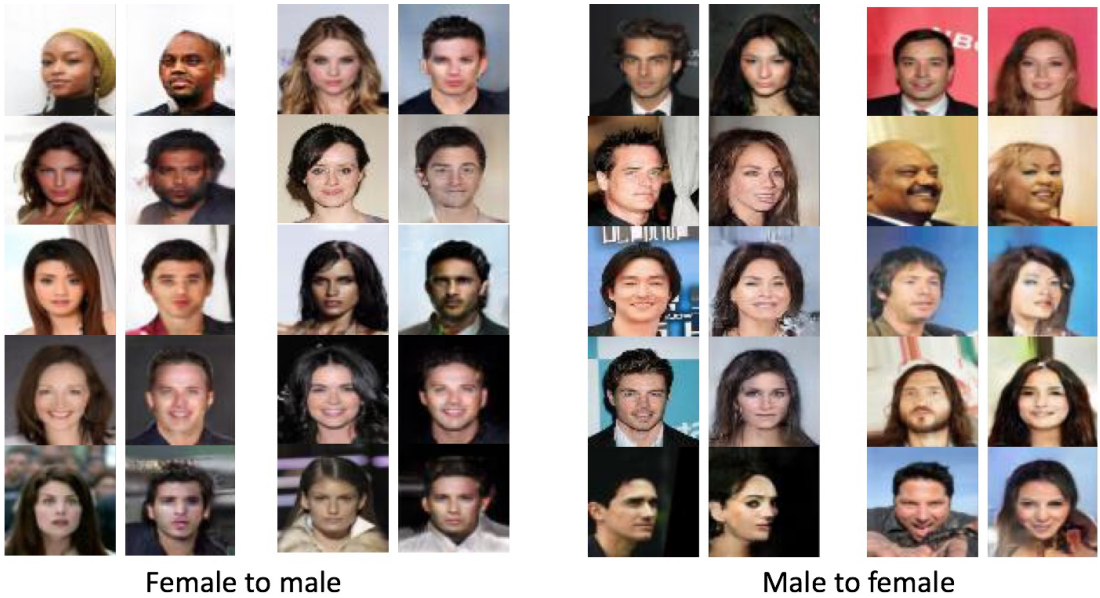

Figure 3. Example results of our method on gender transformation. The odd columns are input images, the even columns are the corresponding synthesized output images.

### 3.3. Translation

After finishing the training of generator and image encoder, as the right of Figure 2 shows, given an image and desired class/domain label, the network firstly maps the image to global latent representation $z$ and then synthesizes the image of target domain by using conditional generator.

## 4. Experiments

In this section, we firstly present results on the celebA face dataset for gender transformation and the presidential debate video for face swapping. The celebA face dataset [11] contains 84434 male images and 118165 female images, which has variety of backgrounds and faces. For the second data set, we detected and extracted face images of Obama and Hillary from two short presidential debate videos, there are 8452 images for Obama and 5065 images for Hillary, which has variety of face expression.

We used the same GAN and image encoder architecture for both datasets. The image size was 64 x 64, with 3 channel colors. We set the number of latent variables $z$ to 100 and embedded the class label to a vector of 5 values. As in auxiliary classifier GAN [15], we used the learning rate of 0.0002, and used the Adam optimizer [8] with momentum of 0.5 for both step 1 and 2. During training, we used batch size of 64, trained for 100 epochs for step 1 and 20,000 steps for step 2. Besides, we randomly flip, zoom in, crop and rotate the training images for data augmentation in step 1. The implementation was built with TensorLayer[2] library on top of TensorFlow [1].

### 4.1. Gender transformation

As Figure 3 illustrated, the proposed method not only learned the characteristics and expressions of human face, but also learn to reconstruct background in some degree. On one hand, the translation has achieved desired performance, with the synthetic image having similar image quality as input images. On the other hand, the network can successfully translate face images for both female to male and male to female, which means, our method is bidirectional.

### 4.2. Face swapping

Figure 4 shows the proposed method learns the expressions and face orientation very efficiently, so it does not violate the context. Such feature is very useful if we want to swap the faces in video.

## 5. Conclusion

In this work, we proposed a simple and effective two step learning method for universal unsupervised image-to-image translation. In next step, we aim to further scale up

| Dataset | Task |
|---|---|
| celebA | gender transformation |
| presidential debate videos | face swapping |

Table 1. Datasets and tasks

[2]https://github.com/zsdonghao/tensorlayer

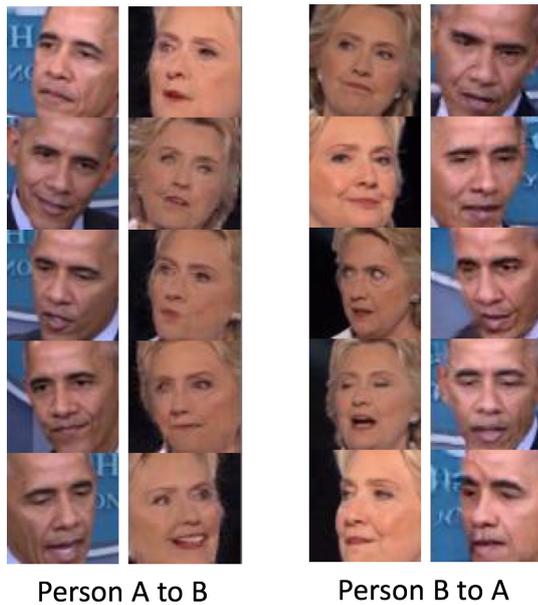

Person A to B     Person B to A

Figure 4. Example results of our method on face swapping

the network for higher resolution and try on more datasets. In the future, it might be interesting to combine symbolic features (like edge detection) with the original data, so that the performance of translation can be further improved. We will also try to make the method more general, to give it better characteristic of one learning algorithm, and apply it to various different other applications. We are considering to proposed an paradigm of "modular learning" which conducts continuous recursive self-improvement with regards to the utility function (reward system). We can view this as second (and higher) order optimization.